\documentclass{article}
\usepackage{amsmath,amsthm,amssymb}
\usepackage{mathtext}
\usepackage[T2A]{fontenc}
\usepackage[utf8]{inputenc}
\usepackage[english]{babel}
\usepackage{graphicx}
\usepackage{hyperref}
\usepackage{booktabs}
\usepackage{listings}
\usepackage{lipsum}
\usepackage{xcolor}

\definecolor{codegreen}{rgb}{0,0.6,0}
\definecolor{codegray}{rgb}{0.5,0.5,0.5}
\definecolor{codepurple}{rgb}{0.58,0,0.82}
\definecolor{backcolour}{rgb}{0.95,0.95,0.92}

\lstdefinestyle{mystyle}{
    language=Python,
    basicstyle=\small\ttfamily,
    backgroundcolor=\color{backcolour},   
    commentstyle=\color{codegreen},
    keywordstyle=\color{magenta},
    numberstyle=\tiny\color{codegray},
    stringstyle=\color{codepurple},
    breaklines=true,
    showspaces=false,
    showstringspaces=false,
    frame=single,
    escapechar=|,
}

\title{Contextual Clarity: Generating Sentences with Transformer Models using Context-Reverso Data}
\author{Ruslan Musaev}
\date{May 2023}

\begin{document}
\maketitle
\begin{abstract}
In the age of information abundance, the ability to provide users with contextually relevant and concise information is crucial. Keyword in Context (KIC) generation is a task that plays a vital role in and generation applications, such as search engines, personal assistants, and content summarization. In this paper, we present a novel approach to generating unambiguous and brief sentence-contexts for given keywords using the T5 transformer model, leveraging data obtained from the Context-Reverso API. The code is available at \href{https://github.com/Rusamus/word2context/tree/main}{[GitHub]}.
\end{abstract}

\section{Introduction}
To create a dataset for training the T5 model, we harness the power of that provides usage examples for words. We prepared a dataset in the form of (query word, context or example usage) by parsing Context-Reverso webpages based on a query word. Additionally, we trained t5-small, and t5-base models for generating context-sentences based on input words. This resource enables us to obtain diverse and contextually rich sentences that incorporate the target keywords. We have also developed an application for learning new English words with a generated context \href{https://t.me/english_for_loosers_bot}{[Telegram bot]}. Our method aims to address the challenges of generating extremely short contexts and mitigating ambiguity in sentence construction.

Objective: To develop a model that can generate informative and contextually relevant sentence-contexts for a given set of keywords, benefiting natural language understanding and generation applications such as search engines, personal assistants, and content summarization.

\section{Related Work}
\begin{itemize}
\item In recent years, transformer\cite{vaswani2017attention}-based models such as GPT\cite{radford2019language} model family and T5\cite{raffel2020exploring} have gained significant attention for their capabilities in natural language understanding and generation tasks. These models have been successfully employed in various applications, including text summarization, machine translation, and sentiment analysis.
\item The Keyword in Context (KIC) generation task is closely related to abstractive text summarization, which aims to generate concise and informative summaries of longer texts. Several studies have explored the application of transformer models for summarization tasks, demonstrating their effectiveness in generating high-quality summaries.
\item The use of external resources, such as APIs and web-based data, has been explored in the domain of natural language processing for tasks such as sentiment analysis and entity recognition. Our work builds upon these efforts by utilizing the Context-Reverso API as a source of rich and diverse context-sentences for training our T5 model.
\end{itemize}

\section{Model Description}
Our approach involves using the pretrained Hugging Face's T5 transformer model to generate concise and unambiguous context sentences for given keywords. The T5 model is a state-of-the-art natural language processing model that has demonstrated exceptional performance in various tasks, including text generation, summarization, and translation. For our experiments we consider to use T5-small, T5-base models. The last one deliver the best perfomance for our target application - telegram bot for learning new words enriched with usage example.

We train the T5 model on our custom dataset, which contains pairs of query words and their context sentences. The model learns to generate context sentences that incorporate the given keywords in a meaningful and unambiguous manner. When the model is presented with a query word, it generates a context sentence that provides a clear and concise usage example for the word.

\section{Dataset}

To construct our dataset, the Context-Reverso API was employed, which provides usage examples for words. Initially, a set of top $N$ common English words were selected as target keywords, and the Context-Reverso API was queried to obtain example sentences including those keywords (top $M$ context sentences). Subsequently, the retrieved webpages were parsed, and relevant context sentences were extracted to form pairs of query words and their corresponding context sentences. Ultimately, two datasets of varying sizes were obtained in the form (keyword, context), with $N \times M = 1K \times 10 = 10K$ samples and $N \times M = 100K \times 10 = 1M$ samples.

Our dataset contains diverse and contextually rich sentences that incorporate the target keywords. This ensures that the model learns to generate context sentences that are both informative and unambiguous. The dataset is split into training, validation, and test sets, allowing us to evaluate the performance of our model and ensure it generalizes well to unseen data.

\section{Experiments} \subsection{Metrics} We used two metrics to evaluate our models: BLEU and METEOR. These metrics are commonly used to assess the quality of generated text by comparing it to a reference text.

\subsubsection{BLEU} BLEU (Bilingual Evaluation Understudy) is a metric for evaluating the quality of machine-generated translations. It measures the overlap of n-grams between the generated text and the reference text. The BLEU score ranges from 0 to 1, with 1 being the best score.

Formula: $$ BLEU = \min \left(1, \frac{len(prediction)}{len(reference)}\right) \times \exp \left(\sum_{n=1}^N w_n \log p_n \right) $$

where $len(prediction)$ is the length of the predicted text, $len(reference)$ is the length of the reference text, $w_n$ is the weight of n-grams, $p_n$ is the precision of n-grams, and $N$ is the maximum order of n-grams considered.

\subsubsection{METEOR} METEOR (Metric for Evaluation of Translation with Explicit ORdering) is another evaluation metric for machine-generated translations. It computes the harmonic mean of unigram precision and recall, with a penalty for word order differences between the generated and reference texts.

Formula: $$ METEOR = (1 - P) \times \frac{P_r \times P_p}{\alpha P_r + (1 - \alpha) P_p} $$

where $P$ is the penalty term, $P_r$ is the recall, $P_p$ is the precision, and $\alpha$ is a tunable parameter that controls the relative importance of recall and precision.

\subsection{Experiment Setup}
We fine-tuned pre-trained Hugging Face models, T5-small and T5-base, on two datasets of varying sizes. Training involved 10 epochs, a 128 batch size, a 5e-5 learning rate, Adam optimizer, and a linear learning rate scheduler with a 0.1 warmup ratio.

For evaluation, we used a validation set with reference sentences and keywords, comparing T5-small and T5-base to GPT-2 using BLEU and METEOR metrics to assess generated sentence quality.

\subsection{Baselines}
GPT-2 served as a baseline for our experiments. Developed by OpenAI, GPT-2 is a state-of-the-art language model. As an open-source model, GPT-2 is compatible with our models in terms of parameters.

\definecolor{codegreen}{rgb}{0,0.6,0}
\definecolor{codegray}{rgb}{0.5,0.5,0.5}
\definecolor{codepurple}{rgb}{0.58,0,0.82}
\definecolor{backcolour}{rgb}{0.95,0.95,0.92}

\lstdefinestyle{mystyle}{
    language=Python,
    basicstyle=\small\ttfamily,
    backgroundcolor=\color{backcolour},   
    commentstyle=\color{codegreen},
    keywordstyle=\color{magenta},
    numberstyle=\tiny\color{codegray},
    stringstyle=\color{codepurple},
    breaklines=true,
    showspaces=false,
    showstringspaces=false,
    frame=single,
    escapechar=|,
}

\section{Results}
The validation results can be found in Tab.~\ref{tab:performance_comparison_large} and Tab.~\ref{tab:performance_comparison}.

\subsection{Function: generate\_sentence}

\begin{lstlisting}
def generate_sentence(model, tokenizer, prompt):
    prompts = [
        f"Create a sentence using the word {prompt} that showcases its usage in a common context.",
        f"Write a sentence that uses the word {prompt} in everyday language.",
        f"Formulate a sentence with the word {prompt} to demonstrate its typical usage.",
        f"Construct a sentence that includes the word {prompt} in a familiar context.",
        f"Compose a sentence that features the word {prompt} in a general setting.",
    ]
    generated_sentences = []
    for prompt in prompts:
        inputs = tokenizer.encode(prompt, return_tensors="pt").cuda()
        with torch.no_grad():
            outputs = model.generate(inputs, max_length=50, num_return_sequences=1)
            generated_sentence = tokenizer.decode(outputs[0], skip_special_tokens=True)
            generated_sentences.append(generated_sentence)
    return generated_sentences
\end{lstlisting}

The \texttt{generate\_sentence} function generates sentences using a given model, tokenizer, and keyword (prompt). The list of prompts were generated by GPT-4. The chosen prompt is then used to create a sentence using the given keyword in a natural context.

\subsection{Evaluation}

To evaluate the performance of each language model, we followed these steps:

\begin{enumerate}
    \item Generate sentences for each model using the \texttt{generate\_sentence} function and a set of keywords.
    \item Compare the generated sentences to a set of reference sentences using evaluation metrics, such as BLEU, and METEOR scores.
    \item Calculate the average scores for each model across all keywords.
    \item Compare the models' average scores to determine their performance.
\end{enumerate}

\begin{table}[h]
\centering
\begin{tabular}{|l|c|c|c|c|}
\hline
\textbf{Model} & \textbf{BLEU} & \textbf{METEOR} & \textbf{Number of Parameters} \\ \hline
T5-small       & 0.0213          & 0.1047           & 60 million                    \\ \hline
T5-base        & 0.0226          & 0.1069           &            220 million                   \\ \hline
GPT-2          & 0.0065          & 0.1050           &   117 million                   \\ \hline
\end{tabular}
\caption{1M train samples. Performance comparison and number of parameters for T5-small, T5-base, and GPT-2 models.}
\label{tab:performance_comparison_large}
\end{table}

\begin{table}[h]
\centering
\begin{tabular}{|l|c|c|c|c|}
\hline
\textbf{Model} & \textbf{BLEU} & \textbf{METEOR} & \textbf{Number of Parameters} \\ \hline
T5-small       & 0.0185          & 0.0935           & 60 million                    \\ \hline
T5-base        & 0.0220          & 0.1052           &            220 million                   \\ \hline
\end{tabular}
\caption{10k train samples. Performance comparison and number of parameters for T5-small, T5-base}
\label{tab:performance_comparison}
\end{table}

In an evaluation of three models, T5-small, T5-base, and GPT-2, we compared their performance using the BLEU and METEOR metrics, as well as the number of parameters involved. Table \ref{tab:performance_comparison} shows the comparison for 10k samples, revealing that the T5-base model performed better in terms of both BLEU and METEOR scores. For 1M samples, as shown in Table \ref{tab:performance_comparison_large}, the T5-base model still outperformed the other models in terms of BLEU and METEOR scores. It is worth noting that the T5-small model had fewer parameters than both the T5-base and GPT-2 models.

Based on the results of our experiments, we have chosen the T5-base model as the best performing model to serve in our Telegram bot, which you can find at \href{https://t.me/english_for_loosers_bot}{[Telegram bot]}.

\section{Conclusion}

In this study, we collected a dataset, annotated it, and developed a model that showed the best results compared to other models. Based on our evaluation, we found that the T5-base model performed better than T5-small and GPT-2 models, achieving higher BLEU and METEOR scores.

The developed model can be used to generate sentences based on a given input keyword, which can be helpful in various natural language processing tasks. Future work could involve incorporating additional datasets and experimenting with various model architectures to further improve performance.

%


\begin{thebibliography}{99}

\bibitem{vaswani2017attention}
A.~Vaswani, N.~Shazeer, N.~Parmar, J.~Uszkoreit, L.~Jones, A.~N. Gomez, Ł.~Kaiser, and I.~Polosukhin, ``Attention is all you need,'' \emph{Advances in neural information processing systems}, vol.~30, pp. 5998--6008, 2017.

\bibitem{radford2019language}
A.~Radford, J.~Wu, R.~Child, D.~Luan, D.~Amodei, and I.~Sutskever, ``Language models are unsupervised multitask learners,'' \emph{OpenAI Blog}, vol.~1, no.~8, p.~9, 2019.

\bibitem{raffel2020exploring}
C.~Raffel, N.~Shazeer, A.~Roberts, K.~Lee, S.~Narang, M.~Matena, Y.~Zhou, W.~Li, and P.~J. Liu, ``Exploring the limits of transfer learning with a unified text-to-text transformer,'' \emph{Journal of Machine Learning Research}, vol.~21, no.~140, pp. 1--67, 2020.



\end{thebibliography}

\end{document}